\documentclass[sigconf]{acmart}
\usepackage{bbding}
\usepackage{multirow}
\usepackage{xcolor}         
\usepackage[ruled,vlined]{algorithm2e}
\definecolor{Mycolor2}{HTML}{202060}

\SetCommentSty{mycommfont}
\usepackage{setspace}
\SetKwInput{KwInput}{Inputs}                
\SetKwInput{KwOutput}{Outputs}              

\usepackage{multirow}
\newcommand{\eg}{\textit{e.g., }}

\AtBeginDocument{%
  \providecommand\BibTeX{{%
    \normalfont B\kern-0.5em{\scshape i\kern-0.25em b}\kern-0.8em\TeX}}}

\setcopyright{acmcopyright}
\copyrightyear{2022}
\acmYear{2022}
\setcopyright{acmcopyright}
\acmConference[MM '22] {Proceedings of the 30th ACM International Conference on Multimedia }{October 10--14, 2022}{Lisbon, Portugal.}
\acmBooktitle{Proceedings of the 30th ACM International Conference on Multimedia (MM '22), October 10--14, 2022, Lisbon, Portugal}
\acmPrice{15.00}
\acmISBN{978-1-4503-9203-7/22/10}
\acmDOI{10.1145/3503161.3548024}

\newcommand{\bb}[1]{\boldsymbol{#1}}
\newcommand{\eat}[1]{}

\begin{document}
\title{Progressive Tree-Structured Prototype Network for End-to-End Image Captioning}


\author{Pengpeng Zeng}
\authornote{Equal Contribution.}
\affiliation{%
  \institution{Center for Future Media \\ University of Electronic Science and Technology of China}
\country{China}
}
\email{is.pengpengzeng@gmail.com}

\author{Jinkuan Zhu}
\authornotemark[1]
\affiliation{%
  \institution{Center for Future Media\\ University of Electronic Science and Technology of China}
  \country{China}
}
\email{jinkuanzhu0@gmail.com}

\author{Jingkuan Song\textsuperscript{\rm 1,2}}
\affiliation{%
  \institution{
  \textsuperscript{1}Center for Future Media\\ University of Electronic Science and Technology of China \\
    \textsuperscript{2}Peng Cheng Laboratory}
    \country{China}
}
\email{jingkuan.song@gmail.com}

\author{Lianli Gao} \authornote{Corresponding author.}
\affiliation{%
  \institution{Center for Future Media\\ University of Electronic Science and Technology of China}
  \country{China}
}
\email{lianli.gao@uestc.edu.cn}

\renewcommand{\shortauthors}{Zeng and Zhu, et al.}

\begin{abstract}
    
    Studies of image captioning are shifting towards a trend of a fully end-to-end paradigm by leveraging powerful visual pre-trained models and transformer-based generation architecture for more flexible model training and faster inference speed. State-of-the-art approaches simply extract isolated concepts or attributes to assist description generation. However, such approaches do not consider the hierarchical semantic structure in the textual domain, which leads to an unpredictable mapping between visual representations and concept words. To this end, we propose a novel Progressive Tree-Structured prototype Network (dubbed PTSN), which is the first attempt to narrow down the scope of prediction words with appropriate semantics by modeling the hierarchical textual semantics. Specifically, we design a novel embedding method called tree-structured prototype, producing a set of hierarchical representative embeddings which capture the hierarchical semantic structure in textual space. To utilize such tree-structured prototypes into visual cognition, we also propose a progressive aggregation module to exploit semantic relationships within the image and prototypes. By applying our PTSN to the end-to-end captioning framework, extensive experiments conducted on MSCOCO dataset show that our method achieves a new state-of-the-art performance with $144.2\%$ (single model) and $146.5\%$ (ensemble of 4 models) CIDEr scores on `Karpathy' split and $141.4\%$ (c5) and $143.9\%$ (c40) CIDEr scores on the official online test server. Trained models and source code have been released at: \emph{\textcolor{blue}{\url{https://github.com/NovaMind-Z/PTSN}}}.
    
\end{abstract}



\maketitle

\section{Introduction}
 \begin{figure}[t]
  \centering
  \includegraphics[width=1.0\linewidth]{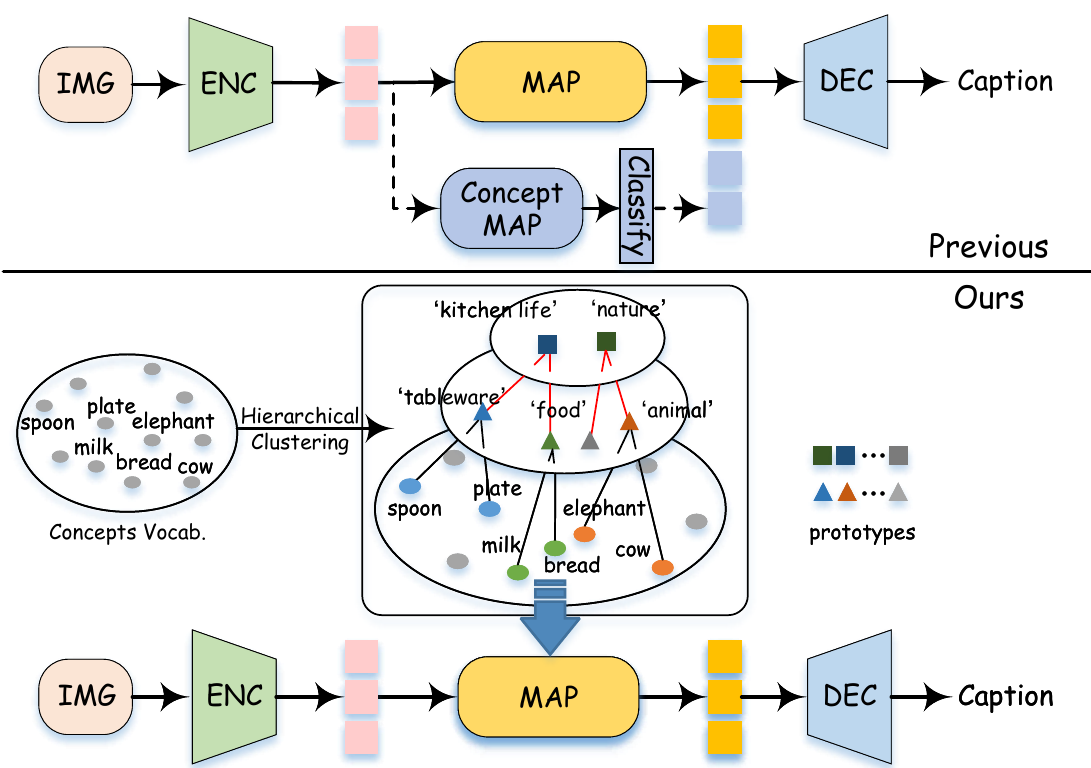}
  \caption{Comparisons between conventional concept-based method and our method for end-to-end image captioning. The upper part depicts the pipeline of previous methods, which fuses the isolated concept embeddings with visual representations. The lower part illustrates our method, which models the hierarchical relationship between all concepts by a set of tree-structured prototypes and progressively guides semantic learning between textual words and visual features.}
  \label{fig1}
\end{figure} 

With the tremendous growth of image materials uploaded to visual online image platforms, \eg Instagram and Twitter, research on automatic image captioning has received increasing attention in recent years \cite{BottomUp,E2ESwin,RST}. Image captioning can lead to substantial practical impacts in various applications, \eg content-based image retrieval and recommendation \cite{contentRetireval,DBLP:journals/tcsv/TianSOS22,zhang2022progressive}, intelligent blind guidance \cite{blind,zhang2020rich,gao2018examine} and human-computer interaction \cite{visualDialog, visualDialog2,zeng2021conceptual}. Despite the significant advances in computer vision and natural language processing in visual analysis and language understanding, image captioning remains an extremely challenging task because of its specific proprieties, such as rich visual information and sophisticated semantics of descriptions. Such a task can be seen as a leap from the recognition to comprehension level.

At the earlier stage, the proposed image captioning methods usually belong to the two-stage methods, which adopt a pre-trained visual encoder to extract features, and reason words based on offline features iteratively. For example, \cite{BottomUp, M2, Xlinear} adopt a pre-trained object detector, usually trained on image cognition tasks, to extract region features, and design different attention mechanisms \cite{M2, Xlinear} to learn inter-modal relationships within the object features. Although achieving promising results, there exists a discrepancy in both data domain and task formulation between these off-the-shelf feature extractors and downstream image captioning. Besides, extracting offline features from images is time-consuming, making it difficult to apply them in real-time image captioning.

Recently, inspired by the successes of transformer-based models in visual understanding tasks, fully transformer-based end-to-end image captioning models \cite{E2EConcept, E2ESwin}, namely one-stage model, have become a new research trend. In practice, such methods directly take the raw image as input into vision transformer \cite{vit} or its variants (\eg Swin Transformer \cite{swin}) for end-to-end training. From Table~\ref{table0}, we can see that compared with the two-stage methods, the one-stage methods achieve an overall considerable improvement of about 5.36\% on average, which shows the benefit of the holistic end-to-end parameter optimization.  

In spite of the extraordinary potential performed in the end-to-end transformer-based methods, how to align their powerful visual features with linguistic words remains to be largely under-explored, which is the core problem for image captioning. Current methods \cite{E2ESwin,E2EConcept} solve this problem by extracting concept or attribute representations from the given image to assist text generation as shown in the upper part of Figure~\ref{fig1}. However, these methods only consider superficial and isolated information of concepts, neglecting the rich structured semantic information in the textual domain, which makes it difficult for visual representations to choose an appropriate word from a wide of candidates. As shown in Figure \ref{fig3}, the concepts `mushrooms' and `cheese' both belong to a parent concept like `pizza ingredients'. Straightly mapping these two detailed concepts with an image of pizza without such structured information is difficult.

\begin{table}[]
\caption{Performance comparison between two-stage methods and one-stage methods.}
\label{table0}
\resizebox{\linewidth}{!}{%
\begin{tabular}{cllc}
\hline
                           & \multicolumn{1}{c}{Method}                                                       & \multicolumn{1}{c}{Backbone} & CIDEr \\ \hline
\multirow{3}{*}{\begin{tabular}[c]{@{}c@{}}Earlier Stage\\ (two-stage model)\end{tabular}} & RSTNet \cite{RST}                                                                          & ResNeXt-101                  & 133.3   \\
& RSTNet \cite{RST}                                                                          & ResNeXt-152                  & 135.6   \\
                           & DLCT \cite{DLCT}                                                                            & ResNeXt-101                  & 133.8   \\ \hline
\multirow{3}{*}{\begin{tabular}[c]{@{}c@{}}Current Trend\\ (one-stage model)\end{tabular}} &  Transformer \cite{E2ESwin}
                       & Swin Transformer              & 136.4   \\
                           & PureT  \cite{E2ESwin}                                                                          & Swin Transformer              & 138.2   \\
                           & Ours                                                                             & Swin Transformer              & 144.2   \\ \hline
\end{tabular}%
}
\end{table}

\par In this paper, we propose a novel Progressive Tree-Structured prototype Network (dubbed PTSN), which progressively guides end-to-end image captioning based on different textual semantic level information. Specifically, we first design a novel concept embedding method, named tree-structured prototype (TSP). As shown in the lower part of Figure~\ref{fig1}, the concept `spoon' and `plate' belong to a semantic prototype `tableware', and the prototypes standing for `food' and `tableware' can be further integrated into a higher-level prototype which represents `kitchen life'. These tree-structured prototypes capture the interdependent and hierarchical relationships between concept words, and these relationships can help our model to generate more fine-grained and reasonable phrases and collocations, such as `spoon and plates', `sing a song', etc. To utilize these prototypes, we propose a progressive aggregation (PA) module for learning the associations between visual features and semantic prototypes from coarse to fine in a top-down manner. Under the guidance of such tree-structured semantics, our final obtained visual feature can own the high-quality semantic information, which leads to a promising quality for captions. Our main contributions can be summarized as follows:
\begin{itemize}
    \item To our best knowledge, we are the first attempt to model the hierarchical semantic information of concepts, named tree-structured prototype (TSP), on the image captioning model. 
    
    \item We propose a progressive aggregation (PA) module to make visual grid features capture such information from the tree-structured prototypes in a coarse-to-fine way.

    \item We conduct extensive experiments on MSCOCO dataset and achieve a new state-of-the-art result, outperforming other competitors by a large margin and even better than many large-scale visual-language pre-training methods.  
\end{itemize}

\eat{The rest of this paper is organized as follows. Related work is introduced in Sec.~\ref{related_work}. In Sec.~\ref{encoder_decoder} and Sec.~\ref{method}, we detail the descriptions of encoder-decoder framework and our proposed method. The experimental results are described in Sec.~\ref{experiments}. Finally, we conclude our work in Sec.~\ref{conclusion}.}

\begin{figure*}
    \centering
    \includegraphics[scale=0.9]{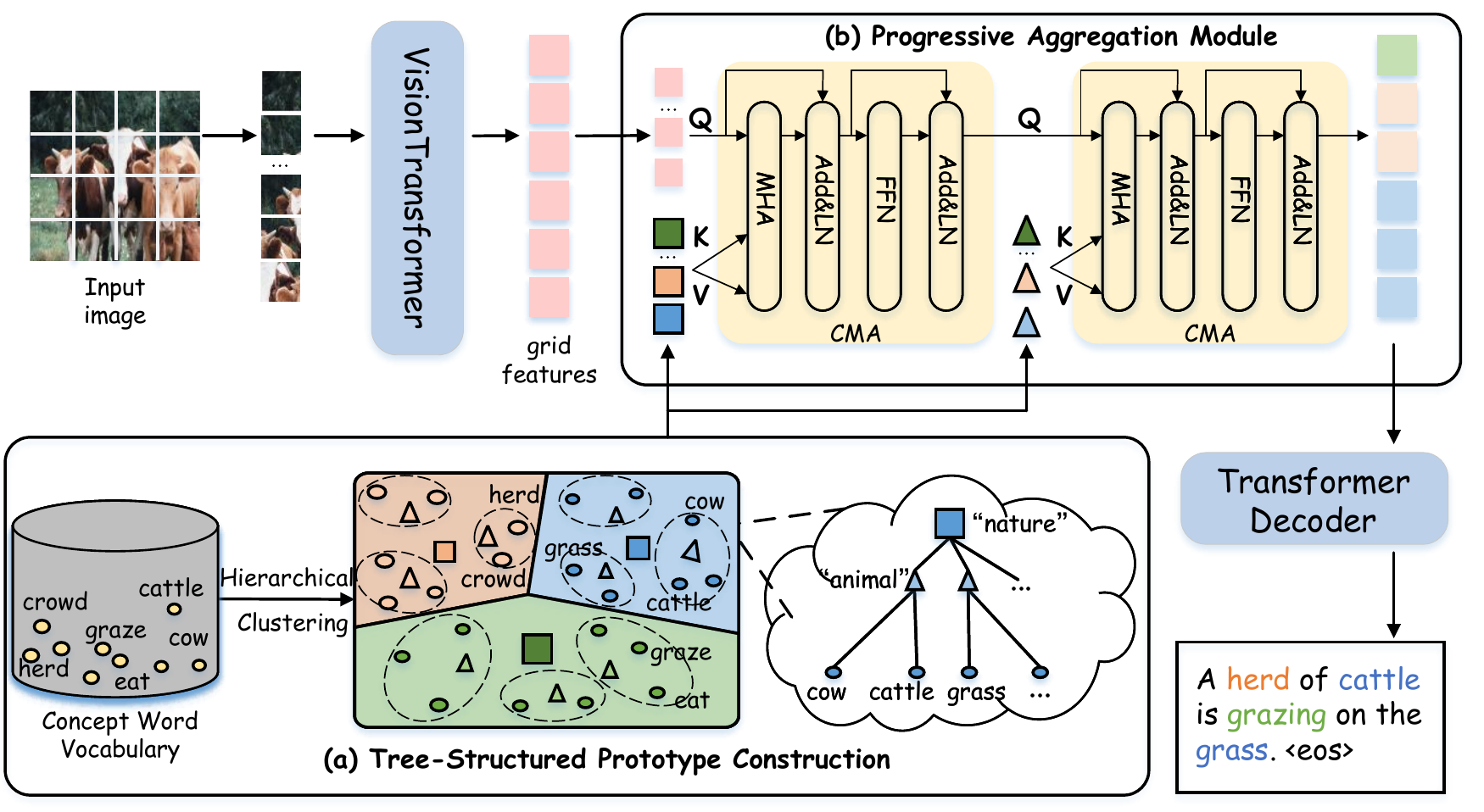}
    \caption{The overview of our proposed framework (PTSN)
    for end-to-end image captioning. It consists of four main components: 1) Vision Transformer, which extracts grid features from the raw image; 2) Tree-Structured Prototype construction module, which builds tree-structured semantic prototypes based on concepts, 3) Progressive Aggregation module, which injects tree-structured prototypes into grid features to obtain semantic-enhanced visual features; and 4) Transformer Decoder for generating final descriptions.}
    \label{fig2}
\end{figure*}

\section{Related Work}
\label{related_work}
\subsection{Image Captioning}
Image captioning is an active research area where excellent works have continued to emerge in the past years. We roughly divide these works into three categories by network architecture. In the first category, CNN-LSTM based methods \cite{ShowTell, showAttendTell, showObserveTell,SemanAtt, bilstm} utilize CNN to extract image features and decode them to words by LSTM. \cite{showAttendTell} introduces an attention-based model which describes the salient objects of the given image. \cite{showObserveTell} proposes an attribute-inference module, which helps the model to observe the image attribute before generating captions. 

\par Afterward, almost all image captioning models \cite{Xlinear, DRT, acm1, acm2, acm3, zeng_s2, bin2021entity} follow a two-stage paradigm and adopt region features (Faster-RCNN) to replace the grid CNN features, which results in a competitive performance. \cite{Xlinear} introduces an attention block to exploit the spatial and channel-wise bi-linear attention distribution to capture the interactions between object features. \cite{DRT} improves the orientation perception between visual features by incorporating the relative direction embedding into multi-head attention. These methods make up the second category but suffer from heavy time costs problem.

\par Recently, end-to-end image captioning methods \cite{E2ESwin, E2EConcept} appear with the development of vision transformers \cite{vit, swin} and achieve significant results. \cite{E2ESwin} first uses a fully-transformer model for image captioning and proposes a refining encoder to refine the grid features by capturing the relationships between them. \cite{E2EConcept} introduces an extra module to predict the semantic concepts and then incorporate them into the visual grids. These third-category methods confirm that vision transformers have the potential to further narrow the gap between visual grids and textual concepts, and how to make these powerful visual grids capture sophisticated internal relationships among concepts is worth exploring.

\subsection{Visual Pre-trained Model}
The development of vision-language (VL) tasks and the visual pre-trained models is inseparable, and the visual pre-trained models provide the representative visual features for VL downstream tasks. Early VL works adopt traditional CNN (e.g. ResNet \cite{resnet}, VGG \cite{vgg}) as the visual feature extractor. These CNNs are pre-trained on image classification datasets such as ImageNet. Then the appearance of Faster-RCNN \cite{fastrcnn} largely influences VL tasks. The Faster-RCNN is trained on the object detection dataset (e.g. VG \cite{VG}) and it offers more fine-grained semantic annotations for each image and processes images with higher resolutions than the aforementioned CNNs (e.g.$448 \times 448$ v.s. $256 \times 256$) for high-quality region features. As for recently proposed vision transformers \cite{vit, swin}, they encode the image as a sequence of visual tokens, using transformer blocks to extract token-level features for images. Depending on the transformer-based structure, vision transformers have larger receptive fields and no inductive bias compared with traditional CNN architecture. Thus they achieve a leading performance in the large-scale visual dataset such as ImageNet-21k. Also, their transformer-based structure is naturally suitable for cooperating with VL downstream tasks.       

\section{Our model structure}
\label{method}

In this section, we present our Progressive Tree-Structured prototype Network (PTSN). The overview of PTSN is depicted in Figure~\ref{fig2}. Technically, the raw image $I$ is first fed into the vision $Encoder$-a vision transformer (VT) to extract the grid feature $\bb{G}$. Then a novel embedding method, tree-structured prototype (TSP), is introduced to take the textual concept $C$ from the word vocabulary as the input to obtain the tree-structured prototypes $\bb{Z}$ in the textual domain. Next, we propose a progressive aggregation (PA) module as our $Map$ to learn the interactions between the visual feature and prototypes and generate semantic-enhanced visual representation $\bb{\hat G}$. Finally, a multimodel decoder is followed to predict the final description $S$. Based on the one-stage framework, our overall pipeline can be summarized as:
\begin{equation}
    \begin{array}{l}
        \bb{G} = VT(I),\\
        \bb{Z} = TSP(C),\\
        \bb{\hat G} = PA(\bb{G},\bb{Z}),\\
        S = Decoder(\bb{\hat G}),
    \end{array}
\end{equation}
where the $Decoder$ is implemented by a vanilla transformer decoder. 

\subsection{Vision Transformer}
\label{VT}
To obtain a representative visual feature for image captioning, we choose the powerful vision transformer as the visual $Encoder$, whose superiority has been proved in many compute vision tasks. In the proposed PTSN, we choose the Swin Transformer as our encoder backbone, which establishes a fully transformer-based framework and enable end-to-end training for image captioning from the raw image.

In practice, for the raw images, we first partition the image $I \in R^{H \times W \times 3}$ into $N$ disjoint patches $I_{p} \in R^{P \times P \times 3}$, where $\{H, W\}$ and $\{P, P\}$ denote the size of the raw image and image patch, respectively. The number of patches $N$ is $(H\times W)/P^2$, which also serves as the effective input sequence length for Swin Transformer. Then, these patches are flattened and mapped into $d$-dimensional vectors via a trainable embedding layer to obtain patch embeddings. For retaining positional information, position embedding is fused into the patch embedding as \cite{swin}. Next, the patch embeddings pass through $4$ stages, where each stage involves a Patch Merging layer and different number of successive Swin Transformer blocks to capture a hierarchical representation. The Patch Merging layer reduces the number of tokens fourfold by concatenating the features of each group of $2\times2$ neighboring patches. And the Swin Transformer block consists of a shifted window-based multi-head self-attention, Multilayer Perceptron, GELU nonlinear layer and layer normalization. To this end, we use the grid features $\bb{G}$ as the final representation from the last stage, which is the input of the Progressive Aggregation module detailed below.

\subsection{Tree-Structured Prototype Construction}
\label{HSP}
To obtain a set of tree-structure prototypes (TSP) for endowing generated captions with fine-grained semantics, we introduce how to construct TSP in this section. As aforementioned, previous approaches process the visual concepts equally, which ignore the hierarchical subordination information and co-occurrence information existing in the textual commonsense. For example, `cow' is a subclass of the `mammal', and `cloud' often appears together with `sky'. This tough problem may result in inappropriate words and inaccurate descriptions in image captioning. Based on the above concerns, we propose a novel tree-structured prototype construction module, which learns the hierarchical semantic structure in the textual space. Besides, our prototypes learn in an implicit way without additional annotations.

To acquire the tree-structured prototypes, we firstly select all verbs, adjectives and nouns from word vocabulary as concept words $C$. A pre-trained word embedding (\eg GLoVe \cite{glove}, BERT \cite{bert} or CLIP \cite{clip}) is used to extract the concept representation $\bb{X}$ for these concept words. Then, we adopt a simple and effective hierarchical clustering algorithm to construct a series of tree-structured prototypes $\bb{Z}$, as described in Algorithm~\ref{Alg.HSPC}. Given the concept representation $\bb{X}$, the number of hierarchies $L$ and the number of prototypes in each hierarchy $F_{1},...,F_{L}$, a clustering algorithm (e.g. \text{k-means} \cite{k-means} or gmm \cite{GMM}) is applied upon $\bb{X}$ to obtain the fine prototypes $\bb{Z_{1}}$ of the first hierarchy. Each prototype represents a center of several concepts with similar meanings, such as `cow' and 'cattle'. After that, we progressively apply the clustering operation to these prototypes to obtain coarser prototypes $\bb{Z_l}$ at higher semantic level, where $l \in (2,L)$. For example, the second-level prototypes $\bb{Z_{2}}$ represent some higher-level topic information like `nature', `sports', `art', etc. More visualizations of the tree-structured prototypes are shown in Section ~\ref{vis}.

\begin{algorithm}[!t]
    \caption{Tree-Structured Prototype Construction.}
    \label{Alg.HSPC}
    \SetAlgoLined
    \SetNoFillComment
    \SetInd{0.5em}{2.5em}
    \KwInput{Concept word representation $\bb{X}$, amount of hierarchies $L$,  numbers of prototype per hierarchy $F_{1},...,F_{L}$}
    \KwOutput{tree-structured prototypes $\bb{Z}= \{ z^{f}_{l} \}$, $(f=1,..,F_{l}, l=1,..,L)$}
    ${\{z^{f}_{1} \}}_{f=1}^{F_{1}} \leftarrow clustering(\bb{X})$
    
    \For{${l = 2}$ to ${L}$}
    {
        ${\{z^{f}_{l} \}}_{f=1}^{F_{l}} \leftarrow clustering({\{z^{f}_{l-1} \}}_{f=1}^{F_{l-1}})$
    }
\end{algorithm}

\subsection{Progressive Aggregation Module}
\label{PA}
To utilize the above tree-structured prototypes for assisting description generation, we propose a progressive aggregation module to aggregate the visual grid features and the prototypes to obtain semantic-enhanced visual features. Specifically, the progressive aggregation first enhances visual grids with coarse prototypes, and then the enhanced grid features utilize fine prototypes for detailed semantic information. By injecting the tree-structured prototypes into visual grid features in such a coarse-to-fine manner, the final refined grid features are able to capture the structured semantic information which is related to the image content. Notably, the progressive aggregation module adopts cross-modal multi-head attention, which is the core block of the PA module.
 
Formally, given grid features $\bb{G}$ and tree-structured prototypes $\bb{Z}$, we send them into sequential Cross-modal Multi-head Attention (CMA) blocks. The CMA block regards visual grid features as query, concept prototypes as key and value, achieving multi-modal interaction to enrich the visual grids with related textual information. The $i$-th CMA block is formulated as follows:  
\begin{equation}  
\label{equ.5}
    \begin{array}{l}
        \bb{\tilde{G}_{i}} = LN(\bb{G_{i}} + MHA(\bb{W_{Q}Z_{i}},\bb{W_{K}Z_{i}},\bb{W_{V}Z_{i}})),\\
        \bb{G_{i+1}} = LN(\bb{\tilde{G}_{i}} + FFN(\bb{\tilde{G}_{i}})),
    \end{array}
\end{equation}

where $MHA$, $LN$, $FFN$ are the vanilla Multi-Head Attention, Layer Normalization, Feed Forward Network introduced in \cite{Transformer}, respectively. $\bb{Z_{i}}$ denotes $i$-th layer prototype. $\bb{W_{Q}}, \bb{W_{K}}, \bb{W_{V}} \in \mathcal{R}^{D\times{D}} $ are learnable parameters and $D$ is the feature dimension. For simplicity, we define the following formulation to represent the above process:
\begin{equation}
\label{equ.6}
    \begin{array}{l}
    \bb{G_{i+1}} = CMA_{i}(\bb{G_{i}}, \bb{Z_i}),\;\; i\!=\!{1,...,L},\\
    \end{array}
\end{equation}
where $\bb{G_{1}} = \bb{G}$ and $L$ denotes the number of stacked CMA block. After being processed by $L$ CMA blocks, the prototype-based grid features $\bb{\hat G} = \bb{G_{L}}$ are obtained, which will be fed into Transformer Decoder to predict the final description.

\subsection{Training Details}
Given the generated sentence $S=\{s_1, s_2, ..., s_T\}$ and the target ground truth $S^*=\{s^*_1, s^*_2, ..., s^*_T\}$, where $T$ is the length of the sentence, we follow the previous works \cite{RST, M2} to train PTSN by two stages: 1) XE stage, where the optimization objective is cross-entropy (XE) loss between the predicted word and ground truth word; 2) RL stage, where the optimization objective is the Reinforcement Learning (RL) reward computed by CIDEr metric.

\par Firstly, we optimize our model with the cross-entropy at XE stage, where the loss function is shown as follows:
\begin{equation}
\label{equ.7}
{L_{CE}}{\rm{ =   - }}\sum\limits_{t = 1}^T {\log ({p_\theta }(s_t^*|s _{1:t - 1}^*))},
\end{equation}
where $\theta$ is the learnable parameters of our model,  $s_{1:T}^*$ denotes the target ground truth sequence.

\par Next, we directly optimize the non-differentiable metric with Self-Critical Sequence Training \cite{SCST} at RL stage, where the loss function is formulated as:
\begin{equation}
\label{equ.8}
{L_{RL}}{\rm{ =   - }}{{\rm{E}}_{s_{1:T}}}{}_{{p_\theta }}[r({s_{1:T}})],
\end{equation}
where the reward $r(\cdot)$ is the CIDEr score.

During RL stage, we use gradient expression in \cite{M2} following \cite{RST} and the mean of rewards is used rather than greedy decoding. Thus, the gradient expression for each sample is formulated as:

\begin{equation}  
\label{equ.9}
    \begin{array}{l}
        b = \frac{1}{k}(\sum\limits_i^k {r({s_i})} ),\\
        {\nabla _\theta }{L_{RL}} \approx  - \frac{1}{k}\sum\limits_{i = 1}^k {((r(s_{1:T}^i) - b))} {\nabla _\theta }\log {p_\theta }(s_{1:T}^i),
    \end{array}
\end{equation}

where $k$ is the number of the sampled sequences, $s_{1:T}^{i}$ is the $i$-th sampled sequence, and $b$ is the mean of the rewards obtained by sampled sequences.

\section{Experiments}
\label{experiments}

\subsection{Experimental Setting}
\textbf{\emph{Datasets and evaluation metrics.}} 
We evaluate our proposed PTSN on the widely used MSCOCO \cite{MSCOCO} dataset. The dataset contains $123,287$ images, and each image is annotated by five human-labeled captions. For a fair comparison, we take the `Karpathy' split on offline evaluation \cite{karpathy}, where $113,287$, $5,000$ and $5,000$ images are used for training, validation and testing, respectively. In addition, MSCOCO also provides an online evaluation platform for further measuring the performance of the model, where there are $40,775$ images without publicly available human-labeled descriptions. Following previous methods, we report the performances of the model with the standard image captioning metrics via using the official released codes, including CIDEr \cite{cider}, BLEU \cite{bleu}, METEOR \cite{meteor}, ROUGE \cite{rouge} and SPICE \cite{spice}.

\textbf{\emph{Backbone.}} In the paper, the Swin Transformer \cite{swin} is chosen as the encoder's backbone to extract grid features from images\eat{, where the backbone consists of M=4 consecutive Swin Transformer stages}. Since Swin Transformer has different model sizes and computation complexities in different configurations, Swin-B and Swin-L are selected as the backbones in our model. The model with Swin-B is conducted in ablation studies to verify the effect of different components. The detailed configurations of the above two backbones are:
\begin{itemize}
    \item Swin-B: resolution = $224\times224$, Size = 88M
    \item Swin-L: resolution = $384\times384$, Size = 197M
\end{itemize}
\noindent where Swin-L is about $2\times$ the model size compared with the Swin-B.

\textbf{\emph{Implementation Details.}} To build word vocabulary, we remove punctuation, convert all words to lower case and keep the words that occur more than 5 times. Besides, we add three special tokens (`<bos>',  `<eos>' and `<pad>') to vocabulary, with a total size of $10,201$ words. For constructing tree-structured prototypes, we set the number of cluster centers as 400, 800 and 2,000 in different layers. If not specifically specified, the feature dimension $D$ of PA and transformer-based decoder is both 512, the number of heads is 8, and the inner dimension of the FFN module is 2,048.

For training, we implement our PTSN method by PyTorch and run on $4$ NVIDIA V100 GPUs in a distributed manner. Specifically, we employ the Adam optimizer to train all models. In XE stage, we set the total epoch and batch size as 20 and 50, respectively. Then we train the model for 30 epochs and the batch size is 10 in RL stage. Following the epoch decay schedule in\cite{RST}, the adjustment process of the lambda learning rate is defined as follows:
\begin{equation}
lambda\_lr=\left\{\begin{array}{rc}
n / 4 \times base\_lr, & n \leq 3, \\
base\_lr, & 3 < n \leq 10, \\
0.2 \times base\_lr, & 10 < n \leq 12, \\
0.2 \times 0.2 \times base\_lr, & otherwise, 
\end{array}\right.
\end{equation}
where $n$ denotes the number of the current epoch. The learning rate $base\_lr$ is set to $4 \times 10^{-5}$ for parameters of backbone and $4 \times 10^{-4}$ for other parameters. For self-critical sequence training, the learning rate is set to a fixed value with $2 \times 10^{-6}$ for backbone parameters and $2 \times 10^{-5}$ for others.  If the CIDEr score drops in 5 consecutive epochs, the training process will stop in RL stage.

\begin{table}[]
\caption{Performance comparison with other state-of-the-art methods on COCO `Karpathy' test split, in single-model setting. All results are reported after the RL optimization stage. Higher is better for all the values in this table.}
\label{Tab:single}
\resizebox{\linewidth}{!}{%
\begin{tabular}{llcccccc}
\hline
\multicolumn{1}{c}{\multirow{2}{*}{Model}} & \multicolumn{1}{c}{\multirow{2}{*}{Backbone}} & \multirow{2}{*}{B@1} & \multirow{2}{*}{B@4} & \multirow{2}{*}{M} & \multirow{2}{*}{R} & \multirow{2}{*}{C} & \multirow{2}{*}{S} \\
\multicolumn{1}{c}{}                       & \multicolumn{1}{c}{}                          &                      &                      &                    &                    &                    &                    \\ \hline
\multicolumn{8}{c}{Two-Stage Models}                                                                                                                                                                                         \\ \hline
AoANet                                     & F-RCNN101                                     & 80.2                 & 38.9                 & 29.2               & 58.8               & 129.8              & 22.4               \\
$M^{2}$ Transformer                             & F-RCNN101                                     & 80.8                 & 39.1                 & 29.2               & 58.6               & 131.2              & 22.6               \\
GET                                        & F-RCNN101                                     & 81.5                 & 39.5                 & 29.3               & 58.9               & 131.6              & 22.8               \\
X-Transformer                              & F-RCNN101                                     & 80.9                 & 39.7                 & 29.5               & 59.1               & 132.8              & 23.4               \\
DRT                                        & F-RCNN101                                     & 81.7                 & 40.4                 & 29.5               & 59.3               & 133.2              & 23.3               \\
RSTNet                                     & ResNeXt101                                    & 81.1                 & 39.3                 & 29.4               & 58.8               & 133.3              & 23.0               \\
DLCT                                       & ResNeXt101                                    & 81.4                 & 39.8                 & 29.4               & 59.1               & 133.8              & 23.0                  \\ 
RSTNet                                     & ResNeXt152                                    & 81.8                 & 40.1                 & 29.8               & 59.5               & 135.6              & 23.3                  \\ \hline
\multicolumn{8}{c}{One-Stage Models}                                                                                                                                                                                        \\ \hline
ViTCAP                                     & ViT-B                                         & -                    & 40.3                 & 29.4               & 59.5               & 133.6              & 23.3               \\
PureT                                      & Swin-L                                       & 82.1                 & 40.9                 & 30.2               & 60.1               & 138.2              & 24.2               \\
PTSN (Ours)                              & ViT-B                                         &81.7                 & 39.3                     & 29.3          &  58.6             & 134.2                   &22.3                    \\
PTSN (Ours)                              & Swin-B                                       & 81.7                 & 39.7                 & 29.5               & 58.6               & 134.7              & 22.4               \\
PTSN (Ours)                              & Swin-L                                       & \textbf{83.6}        & \textbf{41.7}        & \textbf{30.4}      & \textbf{60.2}      & \textbf{144.2}     & 23.7               \\ \hline
\multicolumn{8}{c}{VL Pre-trained Models}                                                                                                                                                                                    \\ \hline
Oscar                                      & ResNeXt152                                    & -                    & 41.7                 & 30.6               & -                  & 140.0              & 24.5               \\
VinVL                                      & ResNeXt152                                    & -                    & 41.0                 & 31.1               & -                  & 140.9              & 25.2               \\
SimVLM                                     & ViT-L                                         & -                    & 40.3                 & 33.4               & -                  & 142.6              & 24.7               \\ \hline
\end{tabular}%
}
\end{table}

\subsection{Performance Comparisons}
\textbf{\emph{Compared methods.}}
In this section, we compare our method with the state-of-the-arts of image captioning to demonstrate the effectiveness of our model. The comparison methods can be roughly divided into three categories: \romannumeral1) Two-stage methods, which adopt offline features directly to reason descriptions, including AoANet \cite{AoANet}, $M^{2}$ Transformer \cite{M2}, GET \cite{GET}, X-Transofmer \cite{Xlinear}, DRT \cite{DRT}, RSTNet \cite{RST} and DLCT \cite{DLCT}; \romannumeral2) One-stage methods, which optimize feature extraction model and caption generation decoder simultaneously, including ViTCAP \cite{E2EConcept} and PureT \cite{E2ESwin}; and \romannumeral3) Vision-and-Language (VL) pre-trained methods, which utilize large-scale data to learning universal multimodal representation, including Oscar \cite{oscar}, VinVL \cite{VinVL} and SimVLM \cite{SimVLM}. Our proposed method PTSN belongs to the second category, one-stage method.

\par To have a fair competition with the methods above, we follow them to conduct experiments under three settings: Single Model, Ensemble Model, and Online Evaluation.
\par \textbf{\emph{Single Model.}} To directly verify the superiority of the PTSN, we first conduct experiments on single model setting. There are three different backbone configurations: ViT-B, Swin-B and Swin-L and the results are shown in Table~\ref{Tab:single}. From this table, we have the following observations: \romannumeral1) compared with the two-stage methods, most one-stage methods achieve superior performance, especially for proposed PTSN. The PTSN (Swin-L) achieves a higher gain than the best counterpart RSTNet (ResNeXt152), particularly improved by 8.6 points on CIDEr. \romannumeral2) Our method belongs to one-stage methods. Compared with the other one-stage methods, our PTSN surpasses all the other approaches in terms of BLEU-1, BLEU-4, METEOR and CIDEr, while slightly worse on ROUGE with respect to PureT. \romannumeral3) Compared with the VL pre-trained works using large-scale multimodal data, our method also achieves superior performance, with an increase of 1.6 points on CIDEr (v.s. SimVLM). Overall, the results demonstrate the promising potential for our PTSN.

\textbf{\emph{Ensemble Model.}} To further explore the potential of our model, we report the results of the ensemble of $4$ models after RL stage. In practice, the ensemble averages the probability distributions of word prediction of multiple models, trained with different random seeds. The results are shown in Table~\ref{Table:ensemble}. From the table, we can observe that our approach exceeds all previous methods on most evaluation metrics. In particular, the CIDEr score of our ensemble model reaches 146.5\% and outperforms the state-of-the-art method by a considerable margin, which achieves advancements of 5.5\% and 9.0\% to PureT and DLCT, respectively.

\textbf{\emph{Online Evaluation.}} For a fair comparison with other competitors, we further adopt an ensemble of four models with two different backbones: Swin-B and Swin-L, and submit the generated captions by our PTSN to the official online test server. The compared results are summarized in Table~\ref{table_online}. Compared to all the other methods, our model with Swin-L still maintains the best performance on all metrics. In terms of c5 and c40 settings on CIDEr, PTSN surpasses the PureT (Swin-L) by 5.4 and 5.6 points, respectively. Notably, our lightweight model with the Swin-B achieves a comparable performance compared to most state-of-the-art methods.

\begin{table}[]
\caption{Performance comparison with other state-of-the-art methods on COCO `Karpathy' test split, using an ensemble of models. All results are reported after the RL optimization stage. Higher is better for all the evaluations in this table.}
\label{Table:ensemble}
\resizebox{\linewidth}{!}{%
\begin{tabular}{llcccccc}
\hline
\multicolumn{1}{c}{} & \multicolumn{1}{c}{Backbone} & B@1           & B@4           & M             & R    & C              & S    \\ \hline
AoANet               & F-RCNN101                    & 81.6          & 40.2          & 29.3          & 59.4 & 132.0          & 22.8 \\
X-Transformer        & F-RCNN101                    & 81.7          & 40.7          & 29.9          & 59.7 & 135.3          & 23.8 \\
$M^{2}$ Transformer        & F-RCNN101                    & 82.0          & 40.5          & 29.7          & 59.5 & 134.5          & 23.5 \\
GET                  & F-RCNN101                    & 82.1          & 40.6          & 29.8          & 59.6 & 135.1          & 23.8 \\
DLCT                 & ResNeXt101                   & 82.2          & 40.8          & 29.9          & 59.8 & 137.5          & 23.3 \\
PureT                & Swin-L                       & 83.4          & 42.1          & 30.4          & 60.8 & 141.0          & 24.3 \\ \hline
PTSN (Ours)         & Swin-L                       & \textbf{84.3} & \textbf{42.4} & \textbf{30.6} & 60.6 & \textbf{146.5} & 24.0 \\ \hline
\end{tabular}%
}
\end{table}

\begin{table*}[]
\caption{Leaderboard of the published state-of-the-art image captioning models on the COCO online testing server. All results are reported after the RL optimization stage. * indicates the large-scale vision-and-language pre-tained methods.}
\label{table_online}
\resizebox{\linewidth}{!}{%
\begin{tabular}{lcccccccccccccc}
\hline
\multirow{2}{*}{Model}    & \multicolumn{2}{c}{B@1}       & \multicolumn{2}{c}{B@2}       & \multicolumn{2}{c}{B@3}       & \multicolumn{2}{c}{B@4}       & \multicolumn{2}{c}{M}    & \multicolumn{2}{c}{R}   & \multicolumn{2}{c}{C}     \\ \cline{2-15} 
                          & c5            & c40           & c5            & c40           & c5            & c40           & c5            & c40           & c5            & c40           & c5            & c40           & c5             & c40            \\ \hline
Up-Down                   & 80.2          & 95.2          & 64.1          & 88.8          & 49.1          & 79.4          & 36.9          & 68.5          & 27.6          & 36.7          & 57.1          & 72.4          & 117.9          & 120.5          \\
RFNet                     & 80.4          & 95.0          & 64.9          & 89.3          & 50.1          & 80.1          & 38.0          & 69.2          & 28.8          & 37.2          & 58.2          & 37.1          & 122.9          & 125.1          \\
GCN-LSTM                  & 80.8          & 95.9          & 65.5          & 89.3          & 50.8          & 80.3          & 38.7          & 69.7          & 28.5          & 37.6          & 58.5          & 73.4          & 125.3          & 126.5          \\
SGAE                      & 81.0          & 95.3          & 65.6          & 89.5          & 50.7          & 80.4          & 38.5          & 69.7          & 28.2          & 37.2          & 58.6          & 73.6          & 123.8          & 126.5          \\
ETA                       & 81.2          & 95.0          & 65.5          & 89.0          & 50.9          & 80.4          & 38.9          & 70.2          & 28.6          & 38.0          & 58.6          & 73.9          & 122.1          & 124.4          \\
AoANet                    & 81.0          & 95.0          & 65.8          & 89.6          & 51.4          & 81.3          & 39.4          & 71.2          & 29.1          & 38.5          & 58.9          & 74.5          & 126.9          & 129.6          \\
$M^{2}$ Transformer             & 81.6          & 96.0          & 66.4          & 90.8          & 51.8          & 82.7          & 39.7          & 72.8          & 29.4          & 39.0          & 59.2          & 74.8          & 129.3          & 132.1          \\
X-Transformer (ResNet-101) & 81.3          & 95.4          & 66.3          & 90.0          & 51.9          & 81.7          & 39.9          & 71.8          & 29.5          & 39.0          & 59.3          & 74.9          & 129.3          & 131.4          \\
X-Transformer (SENet-154)  & 81.9          & 95.7          & 66.9          & 90.5          & 52.4          & 82.5          & 40.3          & 72.4          & 29.6          & 39.2          & 59.5          & 75.0          & 131.1          & 133.5          \\
RSTNet (ResNext101)        & 81.7          & 96.2          & 66.5          & 90.9          & 51.8          & 82.7          & 39.7          & 72.5          & 29.3          & 38.7          & 59.2          & 74.2          & 130.1          & 132.4          \\
RSTNet (ResNext152)        & 82.1          & 96.4          & 67.0          & 91.3          & 52.2          & 83.0          & 40.0          & 73.1          & 29.6          & 39.1          & 59.5          & 74.6          & 131.9          & 134.0          \\
DLCT (ResNext101)        & 82.0          & 96.2          & 66.9          & 91.0          & 52.3          & 83.0          & 40.2          & 73.2          & 29.5          & 39.1          & 59.4          & 74.8          & 131.0          & 133.4          \\
DLCT (ResNext152)        & 82.4          & 96.6          & 67.4          & 91.7          & 52.8          & 83.8          & 40.6          & 74.0          & 29.8         & 39.6          & 59.8          & 75.3          & 133.4          & 135.4          \\
VinVL*                     & 81.9          & 96.9          & 66.9          & 92.4          & 52.6          & 84.7          & 40.4          & 74.9          & 30.6          & 40.8          & 60.4          & 76.8          & 134.7          & 138.7          \\
PureT (Swin-L)                     & 82.8          & 96.5          & 68.1          & 91.8          & 53.6          & 83.9          & 41.4          & 74.1          & 30.1          & 39.9          & 60.4          & 75.9          & 136.0          & 138.3          \\ 
 \hline
Ours (Swin-B)              & 82.1       &96.4   &66.8       &91.3       &52.1       &83.1       &39.9       &73.2       &29.5       &39.1       &59.3       &74.6       &132.2                &134.8                \\
Ours (Swin-L)              & \textbf{84.0} & \textbf{97.5} & \textbf{69.2} & \textbf{93.2} & \textbf{64.5} & \textbf{85.7} & \textbf{42.1} & \textbf{76.1} & \textbf{30.5} & \textbf{40.2} & \textbf{60.4} & \textbf{75.6} & \textbf{141.4} & \textbf{143.9} \\ \hline
\end{tabular}%
}
\end{table*}

\begin{table}[]
\caption{Effect of different word embedding for prototype construction.}
\label{Table:wore_embedding}
\begin{tabular}{lccccc}
\hline
\multicolumn{1}{c}{Word Embedding} & B@1  & B@4  & M    & R    & C     \\ \hline
GloVe                              & 81.3 & 38.6 & 28.7 & 57.9 & 133.3 \\
BERT                               & 81.4 & 39.1 & 29.1 & 58.4 & 133.9 \\
CLIP                               & 81.7 & 39.7 & 29.5 & 58.6 & 134.7 \\ \hline
\end{tabular}
\end{table}

\begin{table}[]
\caption{Effect of different clustering methods.}
\label{Table4}
\begin{tabular}{lccccc}
\hline
\multicolumn{1}{c}{Clustering Method} & B@1  & B@4  & M    & R    & C     \\ \hline
GMM                      & 81.4 & 39.1 & 29.0 & 58.2 & 134.0 \\
\text{k-means}                                     & 81.7 & 39.7 & 29.5 & 58.6 & 134.7 \\ \hline
\end{tabular}
\end{table}

\subsection{Ablation Studies}

The core components of our proposed PTSN are TSP and PA modules. In this section, we perform comprehensive ablation studies to prove the effectiveness of these two modules. All ablation studies are conducted on the backbone of Swin-B and all results are reported after the RL optimization stage. 

\textbf{\emph{Impact of TSP.}}
In this part, we investigate the impact of our TSP module by conducting experiments on different prototype construction methods. Specifically, we study the word embedding methods and clustering methods used in the TSP module.
\par To figure out how different word embedding methods affect the quality of prototypes, we conduct the experiments with three different word embeddings, including GloVe \cite{glove}, BERT \cite{bert} and CLIP \cite{clip}. The results are reported in Table~\ref{Table:wore_embedding}. From the results, we can see that the performance is optimal when using CLIP embedding. Compared with CLIP embedding, the model with GloVe embedding and that with BERT embedding achieve drops of 1.4\% and 0.8\% on CIDEr, respectively. The reason is that CLIP embedding is based on multi-modal pre-training tasks, and it considers not only the visual information but also the semantic meanings of textual descriptions. Therefore, we select CLIP as our word embedding method.

\par To investigate the effect brought by different clustering methods in TSP module, we select two popular clustering methods: \text{k-means} \cite{k-means} and Gasussian Mixture Model (GMM)\cite{GMM}. Table~\ref{Table4} shows the experimental results. We can see that PTSN with \text{k-means} outperforms that with Gasussian Mixture Model (GMM) with an improvement of 0.7\% on CIDEr. Thus we choose \text{k-means} as our clustering method.

\textbf{\emph{Impact of PA.}} 
In this part, we exploit how PA module affects the performance of our proposed method PTSN by conducting experiments on different prototype configurations. Firstly, we experiment with a single hierarchy in the tree-structured prototype module by 400, 800 and 2,000 prototypes according to Algorithm.~\ref{Alg.HSPC}. Furthermore, we use 2 levels of hierarchical clustering structures to attain progressively coarse-to-fine word generation. The specific configurations and experimental results are shown in Table~\ref{Table2}. From the table, we have the following observations:
\begin{itemize}
    \item The baseline without PA performs the worst, but it still achieves comparable results compared to the two-stage methods, as shown in Table~\ref{Tab:single}.
    
    \item Compared with the baseline, the model obtains a better result equipped with a single hierarchy prototype. Moreover, PTSN achieves the best result when the prototype number equals $800$, and it performs worse when the number becomes $2,000$ or $400$. The reason may be that the TSP degrades close to the baseline when the prototype number in the single hierarchy is too large, and it will regard these concepts almost equally. When the prototype number becomes too small, coarser prototypes bring less distinct semantic information and hinder the model to capture more fine-grained semantic words. Therefore, we choose $800$ as the number of coarse prototypes.
    
    \item Under the condition of multiple hierarchies in tree-structured prototype, we ablate two progressive ways to change the granularity of prototypes: coarse-to-fine way (800-2000), and equal-granularity way (800-800). We find that the model in a coarse-to-fine way (800-2000) outperforms the model with single hierarchy prototypes (800) and the model in equal-granularity way (800-800), which demonstrates the advantage of the proposed progressive learning in PA module.
    
    \item As the MCA block number set for each hierarchy in PA increases, the model further achieves a better performance (\eg configuration `800-800-800' > `800-800' > `800', `800-800-2000' > `800-2000' on CIDEr metric). It indicates increasing the learned parameters is beneficial for better capturing the hierarchical knowledge in TSP.
    
\end{itemize}

\begin{table}[]
\caption{Sensitivity analysis on the number of hierarchies and the number of prototypes.}
\label{Table2}
\resizebox{\linewidth}{!}{%
\begin{tabular}{clccccc}
\hline
\multicolumn{2}{c}{Prototype Configuration}                                                                      & B@1  & B@4  & M    & R    & C     \\ \hline
\multicolumn{2}{c}{Baseline w\textbackslash{}o PA}                                                               & 80.7 & 38.4 & 29   & 58.3 & 132.2 \\ \hline
\multicolumn{1}{c|}{\multirow{3}{*}{\begin{tabular}[c]{@{}c@{}}Single-\\ hierarchy\end{tabular}}} & 400          & 81.3 & 38.9 & 28.9 & 58.2 & 132.9 \\
\multicolumn{1}{c|}{}                                                                             & 800          & 81.1 & 38.8 & 28.9 & 58.2 & 133.5 \\
\multicolumn{1}{c|}{}                                                                             & 2000         & 80.9 & 38.5 & 28.8 & 58.2 & 132.7 \\ \hline
\multicolumn{1}{c|}{\multirow{4}{*}{\begin{tabular}[c]{@{}c@{}}Multi-\\ hierarchies\end{tabular}}}  & 800-800      & 81.3 & 39.0 & 29.0 & 58.4 & 133.8 \\
\multicolumn{1}{c|}{}                                                                             & 800-2000     & 81.6 & 39.0 & 29.0 & 58.3 & 134.1 \\ \cline{2-7} 
\multicolumn{1}{c|}{}                                                                             & 800-800-800  & 81.4 & 39.1 & 28.9 & 58.4 & 133.9 \\
\multicolumn{1}{c|}{}                                                                             & 800-800-2000 & 81.7 & 39.7 & 29.5 & 58.6 & 134.7 \\ \hline
\end{tabular}%
}
\end{table}

\begin{figure*}
    \centering
    \includegraphics[scale=0.9]{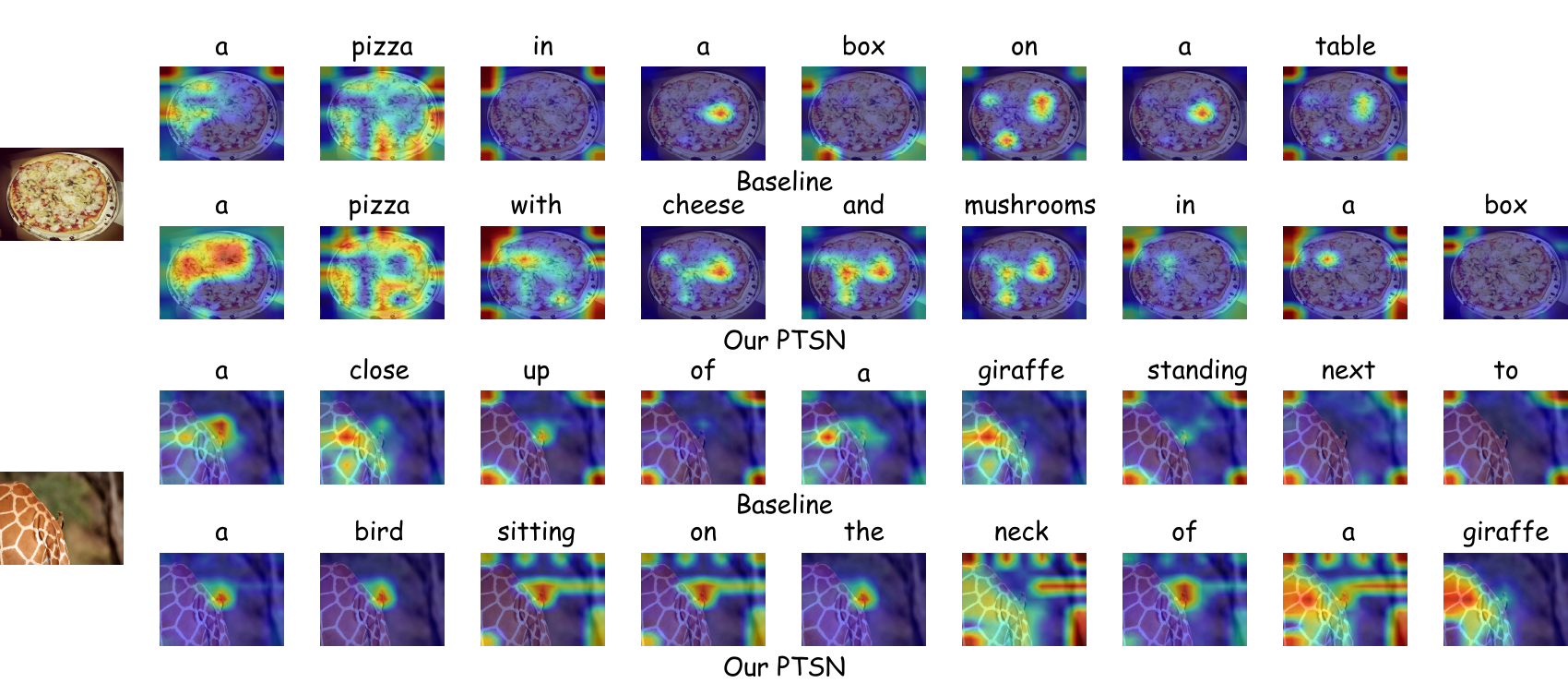}
    \caption{Attention visualization of baseline model and our PTSN. Compared with baseline, our PTSN is able to attend to more fine-grained visual concepts such as `bird', `cheese', and `mushrooms'. }
    \label{fig3}
\end{figure*}

\begin{figure}
    \centering
    \includegraphics[width=0.9\columnwidth]{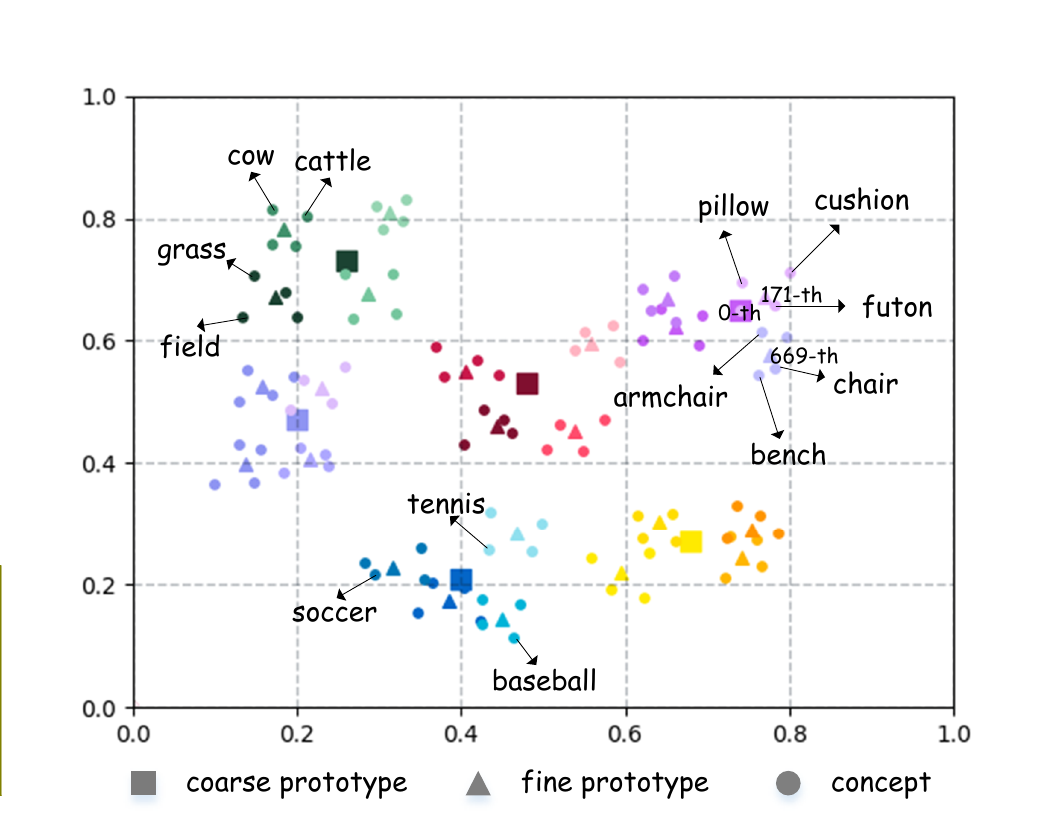}
    \caption{Clustering visualization of tree-structured prototypes. We utilize CLIP to extract concept embeddings and our Algorithm.\ref{Alg.HSPC} to obtain two-hierarchy prototypes. The prototype numbers are $2,000$ and $800$, respectively. }
    \label{fig4}
\end{figure}

\subsection{Visualization}
\label{vis}
In order to better qualitatively evaluate the semantic-enhanced visual representations polished by our PA module, we visualize the importance of the local visual feature contributing to the final output word as in Figure~\ref{fig3}. Technically, we average attention weights of 8 heads in the last multi-head attention layer of the decoder to visualize. We can see that both the baseline model with Swin-B and our PTSN are able to roughly attend to the corresponding grid when generating a word. Remarkably, our model is able to attend to more fine-grained visual concepts like `bird', `cheese', and `mushrooms' in the image.  Specifically, in the first example, both two models detect the salient object `pizza'. However, the PTSN further finds detailed visual content `cheese' and `mushrooms'. This is because these two concepts often co-occur in the real world and are common raw materials of `pizza'. In the second example, we can see that PTSN detects both large-scale visual content `giraffe' and small-scale visual content `bird' while baseline model only notices `giraffe'. It is because both `bird' and `giraffe' are equally important concepts for prototypes, thus our model can describe visual content regardless of their region scales. To conclude, our model successfully models fine-grained visual associations between concepts in the image and generates accurate descriptions.
\par To illustrate the hierarchical semantic structure captured by our TSP module, we select $6$ coarse prototypes and visualize them with their corresponding fine prototypes and concepts. We can observe from Figure~ \ref{fig4} that our tree-structured prototypes successfully describe the internal relationships between concept words. For example, the word `bench', `armchair' and `chair' belong to the $669$-th fine prototype because they all describe the objects used for sitting. The $171$-th fine prototype is also the same, which represents a set of synonyms of `cushion'. Moreover, both the $669$-th and $171$-th fine prototypes belong to the $0$-th coarse prototype, which is believed to represent the concepts describing common furniture.

\section{Conclusion}
\label{conclusion}
In this paper, we propose the PTSN, a novel pure transformer-based model, for end-to-end image captioning. Our proposed tree-structured prototype module obtains the hierarchical semantic structure from previously isolated concepts. Furthermore, we propose a progressive aggregation module to assist visual features to learn more fine-grained visual semantic information from the tree-structured prototypes. Extensive results demonstrate the superiority of our approach that achieves a new state-of-the-art on both offline and online test split. In addition, our method surpasses even some large-scale pre-trained vision-and-language models.

\section*{Acknowledgments}
This work is supported by the National Natural Science Foundation of China (Grant No. 62020106008, No. 62122018, No. 61772116, No. 61872064).

\balance
\bibliographystyle{ACM-Reference-Format}
\bibliography{sample-base}

\end{document}